\documentclass[10pt,twocolumn,letterpaper]{article}

\usepackage{wacv}              

\usepackage[accsupp]{axessibility}
\usepackage{graphicx}
\usepackage{amsmath}
\usepackage{amssymb}
\usepackage{booktabs}
\def\fullname{Zero-Shot VMR\xspace}
\usepackage{bm}
\usepackage{bbm}
\usepackage{multirow}
\usepackage{algorithm,algpseudocode}
\usepackage[numbers]{natbib}
\newcommand{\dz}[1]{{\color{black}#1}}
\newcommand{\yang}[1]{{\color{black}#1}}

\usepackage{adjustbox}

\usepackage[pagebackref,breaklinks,colorlinks]{hyperref}

\usepackage[capitalize]{cleveref}
\crefname{section}{Sec.}{Secs.}
\Crefname{section}{Section}{Sections}
\Crefname{table}{Table}{Tables}
\crefname{table}{Tab.}{Tabs.}

\def\wacvPaperID{0012} 
\def\confName{WACV}
\def\confYear{2024}

\begin{document}

\title{ Zero-Shot Video Moment Retrieval from Frozen Vision-Language Models}

\author{
Dezhao Luo\textsuperscript{\rm 1},
Jiabo Huang\textsuperscript{\rm 1}, 
Shaogang Gong\textsuperscript{\rm 1}, 
Hailin Jin\textsuperscript{\rm 2}, 
and Yang Liu\textsuperscript{\rm 3}\thanks{Corresponding authors} 
\\
\small \textsuperscript{1}{Queen Mary University of London}\\
\tt\small \{dezhao.luo, jiabo.huang, s.gong\}@qmul.ac.uk\\
\small \textsuperscript{2}{Adobe Research}, \small \textsuperscript{3}{WICT, Peking University}\\
\tt\small hljin@adobe.com, \tt\small yangliu@pku.edu.cn
}

\maketitle

\begin{abstract}

Accurate video moment retrieval (VMR) requires universal visual-textual correlations that can handle unknown vocabulary and unseen scenes.
However,
the learned correlations are likely either biased when derived
from a limited amount of moment-text data which is hard to scale up
because of the prohibitive annotation cost (fully-supervised),
or unreliable when only the video-text pairwise relationships
are available without fine-grained temporal annotations (weakly-supervised).
Recently,
the vision-language models (VLM) 
demonstrate a new transfer learning paradigm
to benefit different vision tasks
through the universal visual-textual correlations
derived from large-scale vision-language pairwise web data,
which has also shown benefits to VMR
by fine-tuning in the target domains.

In this work, we propose a zero-shot method for adapting generalisable visual-textual priors from arbitrary VLM to facilitate moment-text alignment, without the need for accessing the VMR data.
To this end, 
we devise a conditional feature refinement module 
to generate boundary-aware visual features conditioned on text queries to enable better moment boundary understanding. 
Additionally, we design a bottom-up proposal generation strategy that mitigates the impact of domain discrepancies and breaks down complex-query retrieval tasks into individual action retrievals, thereby maximizing the benefits of VLM.
Extensive experiments conducted on three VMR benchmark datasets
demonstrate the notable performance advantages of our zero-shot algorithm,
especially in the novel-word and novel-location out-of-distribution setups.
\end{abstract}

\begin{figure}[t]
  \centering
  \includegraphics[width=1\columnwidth]{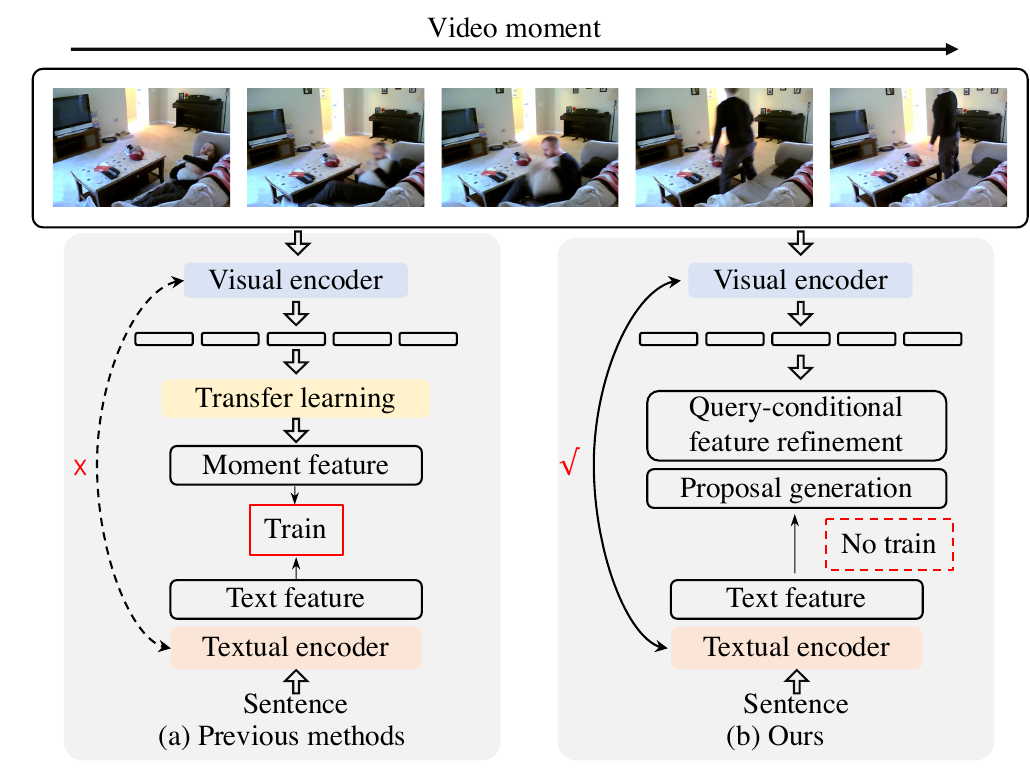}
  \caption{Unlike previous methods~(a) that require additional training to fine-tune pre-trained vision-language models, our method~(b) leverages pre-trained visual-textual alignment to directly predict moment-text alignment, preserving the generality of pre-trained models.
   }
    
  \label{fig:intro}
\end{figure}
\section{Introduction}
\label{sec:intro}

Given a natural video and sentence description, video moment retrieval (VMR) aims to localise a video moment based on the semantics of the sentence. This task 
is challenging as it requires fine-grained moment-text pairs as the learning targets \cite{2dtan,mmn}, which need to annotate not only a sentence but also the temporal position in the video corresponding to the sentence. Since assigning sentences in videos requires high accuracy, it is time-consuming and difficult to scale to web-sized datasets.
To address the problem of lacking fine-grained moment-text pairs for VMR tasks, previous methods \cite{emb,CPL} propose a weakly-supervised setting, which aims to learn the moment-text pairs with weak supervision such as the video-text pairs. To further reduce the reliance on annotations,  recent methods propose an unsupervised VMR to localise the moment with the query chosen from a database \cite{dscnet} or self-generated queries~\cite{psvl,pzvmr}.

Recently, vision-language models (VLM)~\cite{align,clip,furst2021cloob} have shown strong generality in VMR tasks. \dz{Specifically,  PZVMR\cite{pzvmr} and VDI \cite{VDI} have explored  CLIP \cite{clip} by 
fine-tuning the
 pre-learned image-text correlation for moment-text alignment learning}, as shown in Fig. \ref{fig:intro}~(a).  The fine-tuned moment-text alignment from limited video datasets (71K in ActivityNet-Captions \cite{ActivityNet-Caption}) is unlikely to be as generalisable as VLM pre-trained on large-scale  data (e.g. 400M
for CLIP \cite{clip} ).   Different from existing methods, we argue it can bring better generality by directly utilising pre-trained models without any  additional training on domain-specific datasets.  Also, image-level models are less reliable to provide a temporal understanding of the video.  Even though there is still a difficulty in scaling up the fine-grained moment-text annotations, we emphasize the importance of incorporating large-scale video-text models \cite{wang2022internvideo,li2023lavender}.

In this work, we propose a simple yet strong zero-shot VMR that fully satisfies the zero-shot requirement without the need for VMR data access. Our approach relies on the utilisation of the large-scale pre-trained video-text VLM for predicting moment-text alignments, as depicted in Fig.\ref{fig:intro}~(b).
The main challenge lies in the discrepancies between video-text and moment-text domains, as moment-level features necessitate the ability to discriminate between different moments within a video. This means capturing specific temporal information of the moments and understanding their various alignments to a sentence query.
However, the video-text model, originally designed to retrieve textual information for the entire video, struggles to provide accurate temporal boundaries for the target moment.

To address the challenge, 
we adopt snippets as the fundamental units in videos and adapt the video-text model to predict the correlation between snippets and text. 
We recognize that each snippet is more likely to capture short-term actions, so we split the raw-query into multiple simple queries, each containing an individual action that can be better interpreted by a video snippet.
To identify moment boundaries for each simple-query, we propose a conditional feature refinement module to generate boundary-aware features.  Unlike previous methods~\cite{jain2020actionbytes,psvl,pzvmr} which determine boundaries based on abrupt visual changes between snippets, we argue that relying solely on spatial changes is unreliable for reflecting moment changes.
Instead, we propose that the definition of suitable moment boundaries should be conditioned on the query, as different queries may emphasize different visual information.
To reflect moment boundaries based on the query, we refine visual features with their context in a probability indicating how likely they are from the same moment.
By suppressing visual differences within the same moment and enhancing differences between different moments, we generate boundary-aware features that are highly beneficial for VMR, even in cases where precise boundary labels are unavailable.

To generate proposals for the raw-query, we propose a bottom-up proposal generation module. We first cluster the refined snippet features into $k$ proposals for each simple-query. Next, we perform  a Cartesian product operation on the proposals obtained from all simple-queries, enumerating all possible combinations. These combinations are then merged to form final proposals for the  raw-query. The \yang{scores of these proposals} are determined by calculating the average snippet-text correlation \yang{using a }pre-trained VLM.

Our contributions are three-folded: (1) Our zero-shot method eliminates the need for accessing the VMR data by directly applying arbitrary pre-trained VLM for moment-text alignment prediction,  enabling a generalisable VMR without further training. (2) To address the discrepancies between the video-text and moment-text domains, we propose a query-conditional feature refinement module to generate boundary-aware features and a bottom-up
proposal generation module to locate the final moment. (3)~Our method notably outperforms existing unsupervised methods which heavily rely on human-collected videos. Importantly, it also outperforms 
fully-supervised methods when tested on novel-location OOD splits. Furthermore, our experiments reveal that the boundary-aware features have the potential to benefit weakly-supervised VMR where the boundary label is not provided. 
\begin{figure*}[t]
  \centering
  \includegraphics[width=2\columnwidth]{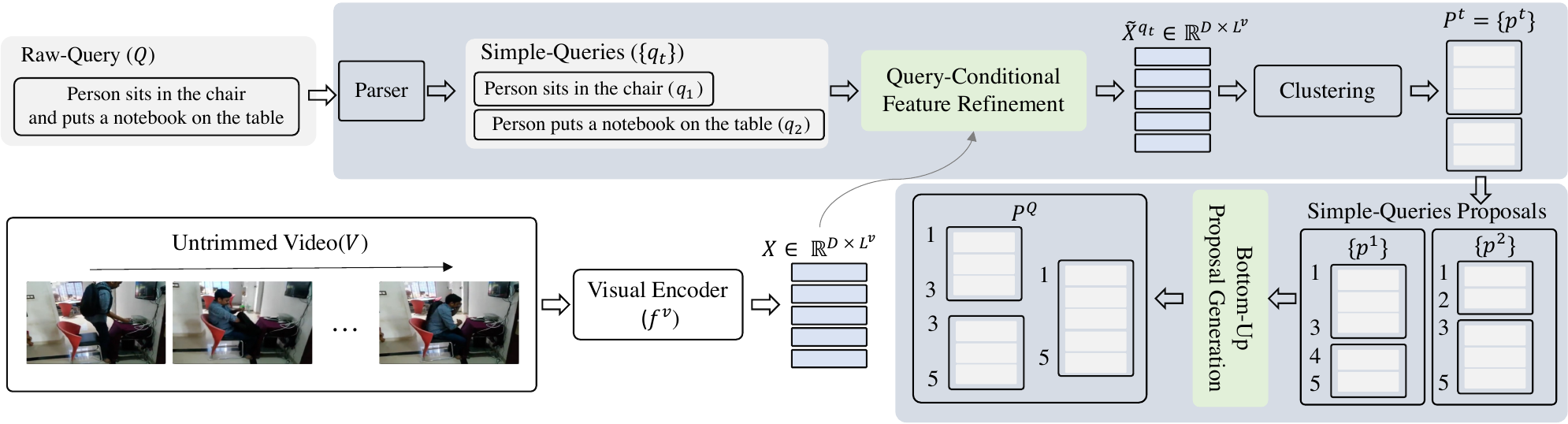}
  \caption{ Our framework.
   We first divide the raw-query $Q$ into multiple simple-queries $\{q_t\}$,  each with one verb.
   Then we cluster the features refined by the query-conditional feature refinement module into proposals as $\{p^t\}$ for each simple-query.  
   In bottom-up proposal generation, we generate the final proposals $P^Q$  for the raw-query by merging all the possible combinations from  its simple-query proposals.
  }
  \label{fig:method}
\end{figure*}

\section{Related Work}

\subsection{Video Moment Retrieval}
Video moment retrieval (VMR) is a challenging task as it requires fine-grained correlation awareness between the video moment and the text. 

For fully-supervised VMR, existing methods \cite{mgsl,vslnet,2dtan,mmn} first generated visual features and textual features from pre-trained models \cite{c3d,DistilBERT}, then they designed a model to align the two modalities.  They inevitably required a large 
number of annotations, which were impractical and unscalable to web-scale datasets.  To alleviate the problem of fine-grained labelling,  weakly-supervised methods \cite{emb,crm} proposed to learn the moment-text alignment with only a given description, relaxing VMR from marking the specific time. 
To further reduce the reliance on human annotations, \citet{pzvmr,psvl} and \citet{dscnet} proposed an unsupervised setting where they generated pseudo queries~\cite{pzvmr,psvl} for the collected videos or chose from a query database \cite{dscnet}. 
To be noted, we regard partially zero-shot methods \cite{psvl,pzvmr} as unsupervised as they still rely on VMR-specific videos, resulting in a suboptimal solution for out-of-distribution (OOD) testing. We argue it is important to reduce the reliance on the collection of VMR data with a strict zero-shot setting for better real-world applications.
 
\subsection{VMR with Vision-Language Pre-Training}
 Large-scale pre-trained vision-language models (VLM) have been explored for better video understanding. To be specific, \citet{wang2021actionclip} proposed to take the class token as an input to the sentence and build video-text alignment from image-text alignment for action recognition. \citet{luo2022clip4clip} proposed to utilise the pre-learned VLM for video retrieval tasks. For VMR, VDI \cite{VDI}  proposed to inject moment information into the pre-trained text encoder and \citet{pzvmr} proposed to utilise the pre-trained image-text alignment as part of the pseudo query. However, existing methods are suboptimal to leverage the generalisable visual-textual alignment learned from large-scale datasets, as they fine-tuned it for moment-text alignment on limited datasets, which is less likely to be as generalisable as the original visual-textual alignments. Moreover, video-level VLM can bring a better temporal understanding compared to image-level models, yet they are not fully explored in previous methods. 

In this work, we propose a zero-shot VMR with the aim of reducing the reliance on VMR-specific dataset collection and directly exploring large-scale video-text pre-trained models for a generalisable VMR.

\section{Methods}
In this section, we first describe the problem and setup of our task and introduce a brief view of vision-language models (VLM), then we
present the details of our approach as well as the rationale behind the design.

\paragraph{Problem Definition.}
\vspace{-\baselineskip}
Given an untrimmed video $V$ that contains $L^v$ non-overlapping snippets $V = \{s_i\}_{i=1}^{L^v}$ and a sentence query $q$, video moment retrieval~(VMR) is performed to locate a video moment $(b^v,e^v)$  according to the semantics of query $q$. The problem can be formulated as:
\begin{equation}
 (b^v,e^v) = {\rm VMR}(f^v(V),f^q(q)),
\end{equation}
where $f^v$ denotes the {\em visual encoder} and its output is $X = \{\bm{x}_i\}_{i=1}^{L^v} \in \mathbb{R}^{D\times L^v}$, $\bm{x}_i$ denotes the visual representation for the $i^{th}$ snippet $s_i$; $f^q$  denotes the {\em textual encoder} that transfers the sentence query $q$ into an embedding. 

\paragraph{Setup.}
\vspace{-\baselineskip}

We detail the difference between our zero-shot setup with existing methods.
Unlike fully-supervised approaches \cite{mmn, 2dtan} that rely on fine-grained moment-text pairs, weakly-supervised methods \cite{emb, CPL} that depend on video-text pairs, unsupervised methods \cite{dscnet}, or partially zero-shot methods \cite{psvl, pzvmr} that rely on human-collected videos, our approach tackles the VMR task without accessing any VMR data, such as videos, queries, or temporal annotations. Our method strictly operates in a zero-shot setting by directly leveraging pre-trained vision-language models, eliminating the requirement for a VMR dataset.

\subsection{Preliminary Study of Vision-Language Models}

\paragraph{Vision-Language Models (VLM).} To learn generalisable and transferable visual and textual models, VLM train a visual encoder (ResNet \cite{resnet} or ViT \cite{vit})  to map high-dimensional images/videos into a low-dimensional
embedding space, and a text encoder (BERT \cite{devlin2018bert}) to generate text
representations from natural language. Then their correlations are learned with a contrastive loss. With a large training vision-text pair dataset (400M in CLIP \cite{clip} and 12M in InternVideo \cite{wang2022internvideo}), they can learn diverse visual-textual correlations that are transferable to downstream tasks.  

\paragraph{Remarks.} 
\vspace{-\baselineskip}

To study if the pre-learned visual-textual alignment is reliable for VMR, we design experiments with image-based CLIP~\cite{clip} and video-based InterVideo \cite{wang2022internvideo} on VMR datasets: Charades-STA \cite{Charades} and ActivityNet-Captions \cite{ActivityNet-Caption}. Firstly, we design a text-retrieval experiment, where the text is the query to retrieve its matched snippet. Given their foreground snippets from the groundtruth, if the VLM can allocate higher scores for the foreground other than the background, we count it as a successful retrieval. The percentage of successful sentence-retrievals out of all the samples is noted as $R^t$ in Table~\ref{table:preli}. Also, we carry out a snippet-retrieval ($R^s$) experiment where the single snippet is the query to retrieve its matched sentences.
In this experiment, we assess whether the VLM can assign higher scores to the matched snippet-text pairs.

As shown in Table~\ref{table:preli}, without pre-learned alignment between I3D \cite{i3d} visual encoder and CLIP~\cite{clip} textual encoder, resulting in random results on both tasks. However, the pre-learned alignments from both CLIP \cite{clip} and InternVideo~\cite{wang2022internvideo} show superior performance, indicating that they are able to understand snippet-query correlations and allocate higher scores for the matched snippet-text pairs.

\begin{table}
\small
  \centering
    \setlength{\tabcolsep}{0.1cm}

  \scalebox{1}{
  \begin{tabular}{cc| cc|cc }
  \hline \multirow{2}{*}{\small{Vision}} & \multirow{2}{*}{\small{Text}} &\multicolumn{2}{c|}{\small{Charades-STA}}   &\multicolumn{2}{c}{
  \small{ActivityNet-Captions}
  } \\
  \cline{3-6}
   &&$R^t$ &$R^s$  &\hspace{10pt} $R^t$  \hspace{10pt}    & $R^s$   \\ 
         \hline

        \small{Random} & \small{Random}& 50.69 &49.67&48.84 & 49.81\\
                 \hline
   \small{I3D} & \small{CLIP}& 51.19 &50.77&47.39& 49.20\\
                 \hline
      \small{CLIP}&\small{CLIP} & 70.43 
      &63.23&69.82& 65.59\\
      \hline
      \small{InternVideo} &  \small{InternVideo} &\textbf{76.09}& \textbf{67.86} &\textbf{71.43}&\textbf{67.60} \\
\hline
  \end{tabular}
  }
  \caption{Preliminary study on the understanding of snippet-text correlation from CLIP \cite{clip} and InternVideo \cite{wang2022internvideo}. The number indicates the retrieval score.  $R^t$ denotes the text-retrieval and $R^s$ the snippet-retrieval task. 
  }
  \label{table:preli}
\end{table}

\subsection{\fullname}
To address the lack of annotations and for a generalisable VMR, we propose a zero-shot method, where we take advantage of the large-scale pre-trained VLM and directly predict moment-text correlations with no additional training on VMR data such as the videos, queries or temporal annotations.  
As shown in Fig.~\ref{fig:method}, we break the task of locating raw-query into multiple simple-query localisations, and we design a feature refinement module to generate boundary-aware features conditioned on each simple-query.

\subsubsection{Query-Conditional Feature Refinement}
\label{Sec:qcpg}

Based on the hypothesis that 
the visual feature undergoes abrupt changes at moment boundaries, previous methods have utilised CNN visual features along with hand-crafted strategies such as k-means \cite{psvl} or a given threshold \cite{jain2020actionbytes, pzvmr} to define moment changes. However, relying solely on visual changes
as indicators
for moment changes is not reliable, as changes in the environment or object appearance may not necessarily correspond to moment transitions.
Considering different queries may focus on different visual information to define a moment, we propose a query-conditional feature refinement module aimed at suppressing the visual differences within a moment and enhancing those between different moments. 
 To be specific, we calculate the probability of a video snippet being in the same moment as its context snippets and refine the visual feature according to the contextual feature.
 
We consider the video as a series of moments, and snippets belonging to the same moment tend to exhibit similar correlation scores to the query describing the moment, while those from different moments show diverging scores. In this regard, we start by  calculating their snippet-query correlation scores by $f^c$: 
\begin{equation}
\label{eq:2}
   f^c(s,q) = {\rm VLM}(s,q),
\end{equation}
where
VLM denotes the pre-trained vision-language model whose input is the snippet $s$ and the query $q$. 

To calculate the probability of a snippet and its context being in the same moment, we first identify the snippet that is most likely to belong to a different moment 
by locating the snippet with the largest snippet-query correlation difference with $s$:
\begin{equation}
\label{eq:3}
\begin{aligned}
D^q &= \{(f^c({s},q)-f^c({s_m},q))^2\}_{m=1} ^{L^v},\\
M^q &= \arg(\max(D^q)),    
\end{aligned}
\end{equation}
where 
$D^q \in \mathbb{R}^{1 \times L^v}$ is the snippet-query correlation difference between $s$  and every snippet  $s_m$ in the video, conditioned on the query $q$; $M^q$ refers to the index for the snippet $s_{M^q}$ which has the largest correlation difference with $s$ and is considered to  belong to a different moment with $s$. As the largest correlation difference captures the maximum disparity between moments, we use it as a metric to compute the probability of two snippets belonging to the same moment:
\begin{equation}
\label{eq:4}
    w^q_{m} =1- \frac{(f^c({s},q)-f^c({s_m},q))^2}{(f^c({s},q)-f^c(s_{M^q},q))^2},
\end{equation}
where $w^q_{m}$ is the probability of $s_m$ being in the same moment with $s$ conditioned on $q$.

 To integrate snippets that  belong to the same moment, which exhibit similar correlation scores to the sentence $q$, we refine the visual feature of $s$ by:
\begin{equation}
\label{eq:5}
   \bm{\tilde{x}}^q= \bm{x} + \lambda \times  X \times W^q \times {\rm mask},
\end{equation}
where $W^q= \{w^q_{m}\}_{m=1}^{L^v}$ and  $\lambda$ is a hyper-parameter; The mask is binary values to filter context snippets. In our implementation, we only consider snippets within a distance of $L^n$ from $s$ and their mask values are set to 1, while snippets outside this range would have a mask value of 0.

\subsubsection{Bottom-Up Proposal Generation}
   
    Since the video-text model has difficulties in generating fine-grained boundaries within the video, we adapt it for snippet-text correlation prediction. As snippets are the fundamental units of a video and are more likely to display short-term actions,  we relax the raw-query retrieval task to multiple simple-query retrievals.
   The motivation behind this approach is to divide complex actions into individual actions to better leverage the video-text model.
   To be specific,  we utilise a language parsing tool to parse the raw-query $Q$ into several simple-queries $q_t$ by extracting the verbs and their corresponding words:
\begin{equation}
\label{eq:6}
 {\rm Parser(Q)}\footnote{Allennlp:https://allenai.org/allennlp}  = \{q_t\}_{t=1}^{L^Q},
\end{equation}
where $L^Q$ is the number of simple-queries extracted from the query $Q$. 
   Then the visual feature $X$ is refined to   $\tilde{X}^{q_t} = \{\bm{\tilde{x}}_i^{q_t}\}_{i=1}^{L^v}$, conditioned on each simple-query $q_t$ with Eq.~\eqref{eq:5}. Then we cluster the features into $k$ proposals:
\begin{equation}
\label{eq:7}
 P^t = {\{p^t_n\}}_{n=1}^{k}, 
\end{equation}
where $P^t$
refers to the proposal list generated for the $t^{th}$ simple-query $q_t$.
   
For proposal scoring,
unlike previous methods 
\cite{pzvmr,VDI}  fine-tuning vision-language models to learn moment-text alignment, 
we simply utilise the  pre-trained video-text  alignment to predict the moment-text alignment with an averaging approach: 
\begin{equation}
\label{eq:8}
    \tilde{C}(p^t) = \frac{\sum_{i=b^{p^t}}^{e^{p^t}}f^c({s_i},{q_t})}{e^{p^t}-b^{p^t}},
\end{equation}
where $\tilde{C}(p^t)$ denotes the  correlation score between proposal $p^t$ and  simple-query $q_t$; $b^{p^t}$/$e^{p^t}$ denotes the beginning and ending snippet index of $p_t$; The snippet-query correlation $f^c$ is calculated as Eq.~\eqref{eq:2}.

 To generate the final proposals for the raw-query $Q$, we propose a bottom-up strategy to merge all the results obtained from simple-queries. To be specific, we first generate proposals $\{P^t\}$ for every simple-queries $\{q_t\}$ using  Eq.~\eqref{eq:7}. Then we use a Cartesian product to enumerate all the possible combinations of these proposals. Finally, we take the union of these proposals from the Cartesian product to generate final proposals and their corresponding scores are averaged from simple-query proposal scores:

\begin{equation}
\label{eq:9}
\begin{split}
    P^Q &= \cup(\{P^{1}  \times \cdots \times P^{L^Q}\}) \\
    &=\{ \{p^{1}  \cup \cdots  \cup p^{L^Q}\}|p^{1} \in P^{1} \land \cdots \land p^{L^Q} \in P^{L^Q}, \\
    & IoU(p^1, \cdots , p^{L^Q})  > 0
    \},\\\
C^Q 
&= \{\{\frac{\sum(\tilde{C}(p^1), \cdots , \tilde{C}(p^{L^Q}))}{L^Q}\}  | \\
&p^{1} \in P^{1} \land \cdots \land p^{L^Q} \in P^{L^Q}, 
     IoU(p^1, \cdots , p^{L^Q})  > 0 \},
\end{split}
\end{equation}
 where ``$\times$" denotes the Cartesian product; $P^Q$ denotes the proposal list for the raw-query $Q$ and $C^Q$ are their corresponding scores. Fig. \ref{fig:method} demonstrates an example of $k$ =2 and ${L^Q}$ = 2. The overall process is summarised in Alg.~\ref{alg}. 

\begin{algorithm}[ht]
\caption{Zero-Shot VMR}\label{alg}
\textbf{Input:} 
  Untrimmed videos $V$,
  A query sentence $Q$,
  A visual $f^v$ encoder,
  A pre-trained model VLM. \\
\textbf{Output:}
  Video moment candidates $(P^Q)$ with their scores~$(C^Q)$. \\
Compute the features of videos $X$ by $f^v$; \\
Generate simple-queries $\{q_t\}_{t=1}^{L^Q}$ (Eq.~\eqref{eq:6}) ; \\
$A \gets \{\}$
  \begin{algorithmic}[1]
\For{$t \gets 1$ to $L^Q$}
{
    \State Calculate the context probability conditioned on $q_t$ as $W^{q_t} = \{w^{q_t}_m\}$, with the VLM (Eq.~\eqref{eq:4}); 
    \State \dz{Generate the boundary-aware features from $X$ to ${\tilde{X}^{q_t}}$ (Eq.~\eqref{eq:5});
    \State Generate proposals $P^t = \{p^t\}$ by clustering the  boundary-aware feature (Eq.~\eqref{eq:7});}
    \State Calculate the correlation score between the proposal and query $\tilde{C}(p^t)$ (Eq.~\eqref{eq:8});
    \State{$A$.{\rm append}($P^t$)}
\EndFor
}

\end{algorithmic}
Unify the Cartesian product of $A$ to generate $P^Q$ and average their corresponding scores as $C^Q$(Eq.~\eqref{eq:9});
\end{algorithm}

\begin{table*}
  \centering
  \setlength{\tabcolsep}{1.8mm}{
  \begin{tabular}{l|c|c|c cc|ccc|ccc|ccc}
    \hline
     \multirow{3}{*}{Method}&\multirow{3}{*}{Year}&\multirow{3}{*}{Setup} &\multicolumn{6}{c|}{Charades-STA}&\multicolumn{6}{c}{ActivityNet-Captions$^\ast$} \\
     \cline{4-15} 
     &&&  \multicolumn{3}{c|}{OOD-1} & \multicolumn{3}{c|}{OOD-2} & \multicolumn{3}{c|}{OOD-1} &\multicolumn{3}{c}{OOD-2} \\
     &&&0.5&0.7 & mIoU&0.5&0.7 & mIoU & 0.5&0.7 &mIoU& 0.5&0.7 &mIoU \\
    \hline

    LGI \cite{lgi2020} & 2020&\multirow{6}{*}{\small{\shortstack{Fully\\-Supervised}}}&42.1& 18.6&41.2& 35.8&13.5&37.1& 16.3 & 6.2 & 22.2&11.0&3.9&17.3 \\
CMI\cite{tsp}&2020&&30.4 & 16.4 &30.3 & 28.1&13.6&29.0&12.3 & 5.2 & 19.1 & 10.0 & 4.2& 16.8 \\
    2D-TAN \cite{2dtan}  & 2020&&27.1 &13.1 &25.7&21.1&8.8&22.5 & 16.4 & 6.6& 23.2 &11.5&3.9&19.4 \\
    DCM\cite{dcm} & 2021 && 44.4 &19.7&42.3&38.5&15.4&39.0&18.2&7.9 & 24.4 & 12.9&4.8 & 20.7 \\
    MMN$^\dagger$ \cite{mmn}& 2022 & & 31.6&13.4& 33.4&27.0&9.3&30.3 &  20.3&7.1&26.2 & 14.1&5.2 & 20.6  \\
   \underline{VDI$^\dagger$ \cite{VDI}} & 2023  &&25.9 &11.9&26.7 &20.8&8.7 &22.0&20.9&7.1 &27.6 &14.3 &5.2&23.7\\
   \hline
      CNM$^\dagger$ \cite{cnm} &2022 &\multirow{2}{*}{\small{\shortstack{Weakly\\-Supervised}}}&9.9&1.7&21.6 & 6.1&0.5&16.6&6.1&0.4&21.0&2.5&0.1&16.8\\
   CPL$^\dagger$ \cite{CPL} &2022& & 29.9&8.5&32.2&24.9&6.3&30.5&4.7&0.4&21.1&2.1&0.2&17.7\\
        \hline
         PSVL$^\dagger$\cite{psvl} &2021&\multirow{2}{*}{\small{\shortstack{Un\\-Supervised}}} &3.0&0.7&8.2&2.2&0.4&6.8&-&-&-&-&-&-\\
         \underline{PZVMR \cite{pzvmr}}  &2022& & -& 8.6 & 25.1 & - & 6.5 & 28.5&-&4.4&\textbf{28.3}&-&2.6&19.1 \\
 \cline{3-3}
 \underline{Ours} &2023&\small{Zero-Shot}&\textbf{40.3}&\textbf{18.2}&\textbf{38.2}&\textbf{38.9}&\textbf{17.0}&\textbf{37.8}&\textbf{18.4}&\textbf{6.8}&21.1&\textbf{18.6}&\textbf{7.4}&\textbf{20.6}\\
 \hline
  \end{tabular}
  }
  \caption{ Novel-location OOD testing.  
  ActivityNet-Captions$^\ast$ denotes the datasets removed from the long-moment for fair comparisons~\cite{dcm}. 
  ``$\dagger$" denotes our implementation with their released models. 
  ``-" denotes the model is not available, and the performance is not reported. Methods using  pre-trained VLM alignments are underlined.  }
  
  \label{table:moment-OOD}
\end{table*}

\section{Experiments}
With the aim of fully exploring the generalisable video-text alignment from large-scale pre-trained models, we propose to directly utilise their visual and textual encoder for video moment retrieval (VMR) without any further training.  To validate the generality and effectiveness of our method, we compare with existing methods on both out-of-distribution (OOD) and independent and identically distributed (IID) data splits.
In this section, we first explain the implementation details and then report our results in comparison with recent methods with a specific focus on unsupervised methods where no annotation is required. Finally, we carry out ablation studies to evaluate each module.

\subsection{Experimental Settings}
\subsubsection{Dataset}
\noindent \textbf{Charades-STA \cite{Charades}} is built upon the Charades 
dataset \cite{sigurdsson2016hollywood} for action recognition and localisation.  \citet{Charades} adapt the
Charades dataset to VMR by collecting
the query annotations. The Charades-STA dataset contains
6,670 videos and involves 16,124 queries.
The average duration of the videos is 30.59 seconds and the moment has an average duration of 8.09 seconds. There are 37 long-moments ($L_{mom}$/$L_{vid}$$\geq$ 0.5) out of 16,124 in this dataset.

\noindent \textbf{ActivityNet-Captions} \cite{ActivityNet-Caption} is collected for video captioning task from ActivityNet \cite{caba2015activitynet} where the videos are associated with 200 activity classes. The ActivityNet-Captions dataset consists of 19,811 videos with 71,957
queries.   The average duration of the videos is around 117.75 seconds and the moment has an average duration of 37.14 seconds. 
For this dataset, there are 15,736 long-moments out of 71,957.

\noindent \textbf{TaCoS} \cite{Tacos} consists of 127 videos from MPIICooking \cite{tacos-from}. It is comprised of
18,818 video-text pairs of cooking activities
in the kitchen annotated by \citet{Tacos}.  
 
 \begin{table}
  \setlength{\tabcolsep}{0.08cm}
  \centering
  \begin{tabular}{l|c|cc| cc}
    \hline
    \multirow{2}{*}{Method}&\multirow{2}{*}{ Setup} &\multicolumn{2}{c|}{Charades-STA} & \multicolumn{2}{c}{\shortstack{ActivityNet-Captions}}   \\
    \cline{3-6}
 &          & \multirow{1}{*}{0.5}&\multirow{1}{*}{0.7} &\hspace{10pt} \multirow{1}{*}{0.5} \hspace{10pt} &\multirow{1}{*}{0.7}  \\
    \hline
    LGI &\multirow{3}{*}{\small{\shortstack{Fully\\-Supervised}}}&26.48&12.47&23.10&9.03 \\  
    VISA && 42.35 & 20.88 &30.14&15.90 \\
    VDI & & 46.47& 28.63 & 32.35&16.02 \\
    \hline
        CNM$^\dagger$&\multirow{2}{*}{\small{\shortstack{Weakly-\\Supervised}}}& 32.52&14.82&23.11&10.21 \\
    CPL$^\dagger$ && 45.90 &22.88  &21.71&9.08 \\
        \hline
  Ours &\small{Zero-Shot}&\textbf{45.04}&\textbf{21.44}  &\textbf{24.57}&\textbf{10.54}\\
        \hline
  \end{tabular}
  \caption{Comparison with methods on the novel-word split \cite{visa}. ``$\dagger$" denotes the same with Table \ref{table:moment-OOD}.}
  \label{table:novel_word}
\end{table}

\subsubsection{Implementation Details}

\textbf{Evaluation Metrics.}
We take “R@n, IoU = µ” and “mIoU”
as the evaluation metrics, which denotes the percentage of queries having at least one result whose intersection over union (IoU) with ground truth is larger
than µ in top-n retrieved moments. “mIoU” is the
average IoU over all testing samples. 
We report the results as n $\in \{1\}$
with µ $\in \{0.1, 0.3, 0.5, 0.7\}$.
 Following DCM \cite{dcm}, we collect ActivityNet-Captions$^\ast$ where long-moments ($L_{mom}$/$L_{vid}$ $\geq$ 0.5) are removed from ActivityNet-Captions.
 
\noindent \textbf{Hyper-Parameters.}  For feature extraction on video snippet, we apply I3D \cite{i3d} on Charades-STA and C3D \cite{c3d} on ActivityNet-Captions. For snippet-text correlation, we apply the pre-trained video-level InternVideo model \cite{wang2022internvideo}.

For feature refinement, the context distance ($L^n$) is set to be 2, the $\lambda$ is 0.5. We select k-means for clustering and the $k$ to be  6. 
We sample the length ($L^v$) of each video as 32.  

\begin{table}
  \setlength{\tabcolsep}{0.06cm}
  \centering
  \begin{tabular}{l|c|cc| cc}
    \hline
    \multirow{2}{*}{Method}&\multirow{2}{*}{ Setup} &\multicolumn{2}{c|}{Charades-STA} & \multicolumn{2}{c}{\shortstack{ActivityNet-Captions}}   \\
    \cline{3-6}
 &          & \multirow{1}{*}{0.5}&\multirow{1}{*}{0.7} &\hspace{10pt} \multirow{1}{*}{0.5} \hspace{10pt} &\multirow{1}{*}{0.7}  \\
    \hline
    DCM &\multirow{2}{*}{\small{\shortstack{Fully\\-Supervised}}}&45.47&22.70&22.32&11.22 \\  
Shuffling &&46.67 &27.08&24.57&13.12 \\
    \hline
        CNM$^\dagger$&\multirow{2}{*}{\small{\shortstack{Weakly-\\Supervised}}}& 30.61&15.23&12.89&4.06 \\
    CPL$^\dagger$ && 41.09 &21.91  &8.47&1.67 \\
        \hline
  Ours& \small{Zero-Shot}&\textbf{40.27}&\textbf{16.27}&\textbf{19.40}&\textbf{7.85}\\
        \hline
  \end{tabular}
  \caption{\dz{Comparison with methods on the novel-distribution split~\cite{shuffle}}. ``$\dagger$"   denotes the same with Table \ref{table:moment-OOD}.}
  \label{table:cd-testing}
\end{table}

\subsection{Comparison with the SOTAs}
As a strict zero-shot VMR method requires no access to VMR data, we focus on comparison with existing unsupervised methods~\cite{dscnet,psvl,pzvmr} in both OOD and IID testing.

\begin{table*}

  \setlength{\tabcolsep}{0.3cm}
  \centering
\begin{adjustbox}{width=\textwidth}
  \begin{tabular}{l|c|c|cccc|cccc}
    \hline
     \multirow{2}{*}{Method}&\multirow{2}{*}{Year}&\multirow{2}{*}{Setup} &\multicolumn{4}{c|}{Charades-STA}&\multicolumn{4}{c}{ActivityNet-Captions} \\
     \cline{4-11}
     &&&0.3&0.5&0.7 & mIoU &0.3 &0.5&0.7 & mIoU \\
     \hline
        2D-TAN\cite{2dtan} & 2020 &\multirow{2}{*}{\small{\shortstack{Fully\\-Supervised}}}& 57.31 & 45.75 & 27.88 &41.05 &60.32 & 43.41 & 25.04 & 42.45 \\
          MMN\cite{mmn} &2022&& 65.43&53.25&31.42&46.46 & 64.48&48.24&29.35&46.61 \\
     \hline
          VCA \cite{vca} &2021&\multirow{4}{*}{\small{\shortstack{Weakly\\-Supervised}}}& 58.58 & 38.13 &19.57&38.49 & 50.45& 31.00 & - & 33.15\\

          CNM \cite{cnm}&2022 && 60.04 & 35.15 & 14.95 & 38.11 & 55.68 & 33.33 &13.29&37.55\\
          CPL \cite{CPL} &2022& & 65.99 & 49.05 & 22.61 &43.23& 55.73 & 31.37 & 13.68& 36.65\\
        \citet{huang2023weakly} 
         &2023&&69.16&52.18&23.94&45.20 &58.07&36.91&-&41.02 \\
          \hline
    PSVL \cite{psvl} &2021&\multirow{5}{*}{\small{\shortstack{Un\\-Supervised}}}&46.47 & 31.29 & 14.17 & 31.24 & 44.74 & 30.08 & 14.74 & 29.62\\
       \citet{gao2021learning}&2021&&46.69 & 20.14 & 8.27 & - & 46.15& 26.38 & 11.64 & -\\
           DSCNet \cite{dscnet} &2022&&44.15 & 28.73& 14.67 & - &47.29&28.16&-&- \\
            \underline{PZVMR \cite{pzvmr}} &2022&&46.83 & 33.21 & 18.51 & 32.62 & 45.73 & 31.26 & \textbf{17.84} & 30.35 \\

               \underline{\citet{language-free}} &2023&&52.95 & 37.24 & 19.33 & 36.05& 47.61& \textbf{32.59} & 15.42 & 31.85 \\
               \cline{3-3}
             \underline{Ours} &2023 &\small{Zero-Shot}& \textbf{56.77} & \textbf{42.93}& \textbf{20.13}& \textbf{37.92} &\textbf{48.28}&27.90&11.57&\textbf{32.37}  \\
           \hline
  \end{tabular}
\end{adjustbox}
  \caption{ IID testing results on Charades-STA and ActivityNet-Captions.  ``-" denotes the same with Table \ref{table:moment-OOD}. Methods using pre-trained VLM alignments are underlined.}
  \label{table:original testing}
\end{table*}

\subsubsection{Novel-Location OOD Testing}
We first carry out experiments on a novel-location OOD scenario, where the location of the moment is different in the training set. Following DCM \cite{dcm}, we add a randomly generated video with $p$ seconds in the beginning of the testing video to modify the location of moment annotations. OOD-1 and OOD-2 refers to $p \in \{ 10,15\}$ for Charades-STA and $p \in \{30,60\}$ for ActivityNet-Captions$^\ast$.
As shown in Table \ref{table:moment-OOD}, we compare with existing methods across different setups.
For Charades-STA, we achieve 40.3\%/18.2\% for OOD-1, 
outperforming unsupervised methods with a significant margin. 
For ActivityNet-Captions, we follow DCM \cite{dcm} to remove long-moments, noted as ActivityNet-Captions$^\ast$, for fair comparisons.  As one can see from Table \ref{table:moment-OOD}, our method 
obtains 18.4\%/6.8\% on OOD-1, 
which reaches the SOTA performance among existing unsupervised methods.

It is worth noting that we demonstrate superior performance on OOD-2 over fully-supervised methods. We argue that models trained with a moment-location biased dataset~(21.87\% long-moments in ActivityNet-Captions) are inferior to be applied to novel-location OOD scenarios, highlighting their limitations in real-world applications.

\vspace{-0.1cm}

\subsubsection{Novel-Word OOD Testing}
To further demonstrate our generality, we carry out testing on the novel-word split released by VISA \cite{visa}, where the testing split contains novel-words not seen in the training.
As shown in Table \ref{table:novel_word}, with novel-word testing, we achieve the performance of 24.57\%/10.54\% for ActivityNet-Captions, outperforming the existing weakly-supervised models \cite{cnm,CPL}.

\subsubsection{Novel-Distribution OOD Testing}

  In Table \ref{table:cd-testing}, we further carry out
testing on the novel-distribution split released by Shuffling \cite{shuffle},
where they shuffle the moment and change the distribution of moments in the testing split. One can see from Table \ref{table:cd-testing}, we achieve the performance of 19.40\%/7.85\% for
ActivityNet-Captions, outperforming existing weakly-supervised models \cite{cnm,CPL}.

\begin{table}
  \setlength{\tabcolsep}{0.2cm}

  \centering
  \begin{tabular}{l|c|ccc}
    \hline
     Method&Setup&0.1& 0.3&0.5\\
 
    \hline
    MCN \cite{mcn} & \multirow{5}{*}{\shortstack{Fully-\\Supervised}}&14.42& - & 5.58 \\
    CTRL \cite{Charades} & &24.32& 18.32&13.30 \\
    QSPN \cite{qspn}& &25.31& 20.15&15.32 \\
    2D-TAN \cite{2dtan}& & 47.59&37.29&25.32\\
    MMN \cite{mmn}&& 51.39&39.24& 26.17\\
    \hline

  Ours&Zero-Shot&\textbf{27.49} &\textbf{11.20} &\textbf{5.57}\\

        \hline
  \end{tabular}
  \caption{Comparison on the original split of TaCoS.}
  \label{table:tacos}
\end{table}

\subsubsection{IID Testing}
To evaluate the effectiveness of our method, we also conduct experiments on IID testing, where the training and testing split share independent and identical distribution. 
As shown in Table \ref{table:original testing}, we outperform unsupervised methods \cite{language-free,psvl,pzvmr} on Charades-STA, whilst we don't require any training or access to VMR dataset. For ActivityNet-Captions, we argue existing methods benefit from the moment-location bias on this dataset. Moreover, we carry out experiments on TaCoS in Table~\ref{table:tacos}.

\begin{table}
  \setlength{\tabcolsep}{0.08cm}
  \centering
  \begin{tabular}{ll|cc| cc}
    \hline
    \multirow{2}{*}{Method}& \multirow{2}{*}{VLM}&\multicolumn{2}{c|}{Charades-STA} & \multicolumn{2}{c}{\shortstack{ActivityNet-Captions$^\ast$}}   \\
    \cline{3-6}
         & & \multirow{1}{*}{0.5}&\multirow{1}{*}{0.7} &\hspace{10pt} \multirow{1}{*}{0.5} \hspace{10pt} &\multirow{1}{*}{0.7}  \\
    \hline
    PZVMR&CLIP &-&6.5&-&2.6 \\  
    \hline
    Ours&CLIP &25.9&9.9&15.7&5.3 \\  
    Ours&InterVideo &\textbf{38.9}&\textbf{17.0}&\textbf{18.6}&\textbf{7.4} \\  
    
        \hline
  \end{tabular}
  \caption{Comparison between methods using pre-trained visual-textual alignments from different VLM.}
  \label{table:dif-VLM}
\end{table}

\subsection{Ablation Study}
 We report ablations
 on Charades-STA and ActivityNet-Captions  
 with novel-location OOD-2 testing.

\begin{table}
  \setlength{\tabcolsep}{0.12cm}
  \centering
  \begin{tabular}{l|ccc|ccc}
    \hline
      \multirow{2}{*}{Method} &\multicolumn{3}{c|}{Charades-STA} & \multicolumn{3}{c}{\shortstack{ActivityNet-Captions}} \\
      \cline{2-7}
      & 0.3& 0.5 & 0.7 &0.3& 0.5  & 0.7 \\
     \hline
     w/o QC-FR&58.33&37.28&14.11  & 39.63&24.29&9.70 \\
    w/o BU-PG  &60.01&\textbf{39.01}&16.68 & 38.95 & 23.83&9.86\\
     \hline
     Ours & \textbf{60.22}&38.92 & \textbf{16.96}&\textbf{40.92}&\textbf{25.70}&\textbf{10.56} \\
        \hline
 
        \hline
  \end{tabular}
  \caption{Ablation study of query-conditional feature refinement (QC-FR) and bottom-up proposal generation (BU-PG).}
  \label{table:ablation-compo}
\end{table}

\begin{table}
  \setlength{\tabcolsep}{0.35cm}

  \centering
  \begin{tabular}{l|cccc}
    \hline
     Method& 0.3&0.5&0.7&mIoU\\
 
    \hline
   w/o QC-FR &65.99&49.05&22.61&43.23 \\
   w QC-FR & \textbf{66.91}&\textbf{50.85}&\textbf{24.00} & \textbf{44.00} \\
        \hline
  \end{tabular}
  \caption{Comparison of weakly-supervised CPL \cite{CPL} performance with and without our QC-FR module on Charades-STA.}
  \label{table:cpl}
\end{table}

\dz{\subsubsection{Vision-Language Model Ablation}
In this subsection, we compare the option of VLM between CLIP \cite{clip} and InterVideo \cite{wang2022internvideo}. As shown in Table~\ref{table:dif-VLM}, our proposed method demonstrates better generality from the pre-trained CLIP than previous PZVMR \cite{pzvmr}. Furthermore, with a 
 better understanding of the temporal information from InterVideo, our method achieves further improvement. }

\begin{table}
  \setlength{\tabcolsep}{0.15cm}

  \centering
  \begin{tabular}{l|lll|lll}
    \hline
      \multirow{2}{*}{Method} &\multicolumn{3}{c|}{$L^Q = $ 1 } & \multicolumn{3}{c}{$L^Q$ $\geq$ 2 } \\
      \cline{2-7}
      & 0.3&0.5&0.7& 0.3& 0.5&0.7 \\
     \hline

w/o BU-PG &\textbf{32.80} &\textbf{18.59}&7.79&43.02&27.18&11.02 \\
w/ BU-PG &32.63&18.58&\textbf{7.82}&\textbf{45.85}&\textbf{29.36}&\textbf{12.16}  \\
        \hline

  \end{tabular}
  \caption{Performance on ActivityNet-Captions for raw-queries with different numbers of simple-queries $L^Q$.}
  \label{table:ablation-number}
\end{table}

\subsubsection{Component Ablation}
In this subsection, we evaluate the effectiveness of our components.  As shown in Table \ref{table:ablation-compo}, without our proposed query-conditional feature refinement (QC-FR) module, there is a performance drop on both datasets.  \dz{Also, to validate the advantages of the generated boundary-aware feature in scenarios where boundary labels are absent, we apply the query-conditioned feature refinement module (QC-FR) in weakly-supervised CPL \cite{CPL} on IID testing. One can see from Table~\ref{table:cpl}, by refining the feature with QC-FR, we observe performance gains of 1.80\% and 1.39\% on IoU = 0.5/0.7. 
}

 For bottom-up proposal generation (BU-PG), our method shows enhanced performance when evaluated with ActivityNet-Captions in Table \ref{table:ablation-compo}. 
To further demonstrate the benefits of BU-PG for complex-queries, we present the performance based on the number of simple-queries for each raw-query ($L^Q$). It can be observed in Table~\ref{table:ablation-number} that BU-PG achieves superior performance when dealing with complex-queries that consist of more simple-queries ($L^Q \geq 2$).

\subsubsection{Hyper-Parameter Ablation}
For hyper-parameters, we report the ablation on Charades-STA.
 We ablate the option of clustering method and select k-means as shown in Table \ref{table:ablation-clustering}. Then 
we take 6 as the value of ``$k$" for k-means as shown in Table \ref{table:ablation-kmeans}. The ablations of  ``$\lambda$" and ``$L^n$" for feature refinement are shown in Table~\ref{table:ablation-kernel}.

\begin{table}
  \setlength{\tabcolsep}{0.3cm}
  \centering
  \begin{tabular}{l|cccc
  }
    \hline
      Method 

      & 0.3& 0.5 & 0.7 &mIoU
      \\
          \hline

     Random &45.03&20.37&6.75&27.35 
     \\
\small{Sliding Window} &49.54&38.41&14.62&33.23  \\
     \small{Abrupt Change} & 20.97& 9.95&2.80&14.11 
     \\
     \textbf{\small{K-Means (Ours)}} & \textbf{60.22}&\textbf{38.92} & \textbf{16.96}& \textbf{37.80} \\
        \hline
 
        \hline
  \end{tabular}
  \caption{Ablation study of clustering method on Charades-STA.}
  \label{table:ablation-clustering}
\end{table}

\begin{table}

  \setlength{\tabcolsep}{0.5cm}
  \centering
  \begin{tabular}{l|llll
  }
    \hline
      $k$ & 0.3& 0.5 &0.7 &mIoU \\
     \hline
     5& 59.35& 35.51 & 13.17 &36.47
     \\
     \textbf{6} &\textbf{60.22}&38.92&\textbf{16.96}  & \textbf{37.80}
     \\
     7 & 58.39& \textbf{39.01} & 16.83 & 37.65
     \\
        \hline
  \end{tabular}
  \caption{Ablation study of k-means on Charades-STA.  }
  \label{table:ablation-kmeans}
\end{table}

\begin{table}

  \setlength{\tabcolsep}{0.35cm}
  \centering
  \begin{tabular}{l|l|llll}
    \hline
      $\lambda$ &$L^n$
      & 0.3& 0.5 &0.7 & mIoU  \\
     \hline
     0.1 &\multirow{3}{*}{\textbf{2}} &58.73&38.39&15.75& 37.49 \\
               \textbf{0.5} &&\textbf{60.22}&38.92&\textbf{16.96} & \textbf{37.80} \\

     1& &59.52&\textbf{39.06}&16.48&37.19 \\
     \hline
     \multirow{2}{*}{0.5}&1&59.81 & 38.09& 16.75& 37.65 \\
          &3&59.68&38.55&16.86&37.48 \\
        \hline

  \end{tabular}
  \caption{Ablation study of $\lambda$ and $L^n$ on Charades-STA.}
  \label{table:ablation-kernel}
\end{table}

\section{Conclusion and Future Work}
In this work, we approach the video moment retrieval~(VMR) task by adapting the generalisable video-text pre-trained models without requiring additional training on the target domain. To address the discrepancy between video-text and moment-text domains, we propose a query-conditional proposal generation module to generate boundary-aware features and a bottom-up proposal generation module for complex-query localisation. Superior performances on OOD testing demonstrate our method can extract generalisable moment-text alignments from pre-trained video-text alignments. For future work, one important direction is to address the challenge of understanding the temporal relationship between individual actions which is not captured by video-text pre-training models. 
\section*{Acknowledgements}
This work was supported by the China Scholarship Council, the Alan Turing Institute Turing Fellowship, Veritone, Adobe Research and Zhejiang Lab (NO. 2022NB0AB05).

{\small
\bibliographystyle{plainnat}
\bibliography{egbib}

\begin{thebibliography}{43}
\providecommand{\natexlab}[1]{#1}
\providecommand{\url}[1]{\texttt{#1}}
\expandafter\ifx\csname urlstyle\endcsname\relax
  \providecommand{\doi}[1]{doi: #1}\else
  \providecommand{\doi}{doi: \begingroup \urlstyle{rm}\Url}\fi

\bibitem[Anne~Hendricks et~al.(2017)Anne~Hendricks, Wang, Shechtman, Sivic,
  Darrell, and Russell]{mcn}
Lisa Anne~Hendricks, Oliver Wang, Eli Shechtman, Josef Sivic, Trevor Darrell,
  and Bryan Russell.
\newblock Localizing moments in video with natural language.
\newblock In \emph{ICCV}, pages 5803--5812, 2017.

\bibitem[Caba~Heilbron et~al.(2015)Caba~Heilbron, Escorcia, Ghanem, and
  Carlos~Niebles]{caba2015activitynet}
Fabian Caba~Heilbron, Victor Escorcia, Bernard Ghanem, and Juan Carlos~Niebles.
\newblock Activitynet: A large-scale video benchmark for human activity
  understanding.
\newblock In \emph{CVPR}, pages 961--970, 2015.

\bibitem[Carreira and Zisserman(2017)]{i3d}
Joao Carreira and Andrew Zisserman.
\newblock Quo vadis, action recognition? a new model and the kinetics dataset.
\newblock In \emph{proceedings of the IEEE Conference on Computer Vision and
  Pattern Recognition}, pages 6299--6308, 2017.

\bibitem[Devlin et~al.(2019)Devlin, Chang, Lee, and Toutanova]{devlin2018bert}
Jacob Devlin, Ming-Wei Chang, Kenton Lee, and Kristina Toutanova.
\newblock Bert: Pre-training of deep bidirectional transformers for language
  understanding.
\newblock In \emph{Proceedings of the 2019 conference of the north american
  chapter of the association for computational linguistics: Human language
  technologies, Volume 1}, pages 4171--4186, 2019.

\bibitem[Dosovitskiy et~al.(2020)Dosovitskiy, Beyer, Kolesnikov, Weissenborn,
  Zhai, Unterthiner, Dehghani, Minderer, Heigold, Gelly, et~al.]{vit}
Alexey Dosovitskiy, Lucas Beyer, Alexander Kolesnikov, Dirk Weissenborn,
  Xiaohua Zhai, Thomas Unterthiner, Mostafa Dehghani, Matthias Minderer, Georg
  Heigold, Sylvain Gelly, et~al.
\newblock An image is worth 16x16 words: Transformers for image recognition at
  scale.
\newblock \emph{arXiv preprint arXiv:2010.11929}, 2020.

\bibitem[F{\"u}rst et~al.(2021)F{\"u}rst, Rumetshofer, Tran, Ramsauer, Tang,
  Lehner, Kreil, Kopp, Klambauer, Bitto-Nemling, et~al.]{furst2021cloob}
Andreas F{\"u}rst, Elisabeth Rumetshofer, Viet Tran, Hubert Ramsauer, Fei Tang,
  Johannes Lehner, David Kreil, Michael Kopp, G{\"u}nter Klambauer, Angela
  Bitto-Nemling, et~al.
\newblock Cloob: Modern hopfield networks with infoloob outperform clip.
\newblock \emph{arXiv preprint arXiv:2110.11316}, 2021.

\bibitem[Gao et~al.(2017)Gao, Sun, Yang, and Nevatia]{Charades}
Jiyang Gao, Chen Sun, Zhenheng Yang, and Ram Nevatia.
\newblock Tall: Temporal activity localization via language query.
\newblock In \emph{ICCV}, pages 5267--5275, 2017.

\bibitem[Gao and Xu(2021)]{gao2021learning}
Junyu Gao and Changsheng Xu.
\newblock Learning video moment retrieval without a single annotated video.
\newblock \emph{IEEE TCSVT}, 32\penalty0 (3):\penalty0 1646--1657, 2021.

\bibitem[Hao et~al.(2022)Hao, Sun, Ren, Wang, Qi, and Liao]{shuffle}
Jiachang Hao, Haifeng Sun, Pengfei Ren, Jingyu Wang, Qi~Qi, and Jianxin Liao.
\newblock Can shuffling video benefit temporal bias problem: A novel training
  framework for temporal grounding.
\newblock In \emph{ECCV}, pages 130--147. Springer, 2022.

\bibitem[He et~al.(2016)He, Zhang, Ren, and Sun]{resnet}
Kaiming He, Xiangyu Zhang, Shaoqing Ren, and Jian Sun.
\newblock Deep residual learning for image recognition.
\newblock In \emph{CVPR}, pages 770--778, 2016.

\bibitem[Huang et~al.(2021)Huang, Liu, Gong, and Jin]{crm}
Jiabo Huang, Yang Liu, Shaogang Gong, and Hailin Jin.
\newblock Cross-sentence temporal and semantic relations in video activity
  localisation.
\newblock In \emph{Proceedings of the IEEE/CVF International Conference on
  Computer Vision}, pages 7199--7208, 2021.

\bibitem[Huang et~al.(2022)Huang, Jin, Gong, and Liu]{emb}
Jiabo Huang, Hailin Jin, Shaogang Gong, and Yang Liu.
\newblock Video activity localisation with uncertainties in temporal boundary.
\newblock In \emph{Computer Vision--ECCV 2022: 17th European Conference, Tel
  Aviv, Israel, October 23--27, 2022, Proceedings, Part XXXIV}, pages 724--740.
  Springer, 2022.

\bibitem[Huang et~al.(2023)Huang, Yang, and Sato]{huang2023weakly}
Yifei Huang, Lijin Yang, and Yoichi Sato.
\newblock Weakly supervised temporal sentence grounding with uncertainty-guided
  self-training.
\newblock In \emph{CVPR}, pages 18908--18918, 2023.

\bibitem[Jain et~al.(2020)Jain, Ghodrati, and Snoek]{jain2020actionbytes}
Mihir Jain, Amir Ghodrati, and Cees~GM Snoek.
\newblock Actionbytes: Learning from trimmed videos to localize actions.
\newblock In \emph{Proceedings of the IEEE/CVF Conference on Computer Vision
  and Pattern Recognition}, pages 1171--1180, 2020.

\bibitem[Jia et~al.(2021)Jia, Yang, Xia, Chen, Parekh, Pham, Le, Sung, Li, and
  Duerig]{align}
Chao Jia, Yinfei Yang, Ye~Xia, Yi-Ting Chen, Zarana Parekh, Hieu Pham, Quoc Le,
  Yun-Hsuan Sung, Zhen Li, and Tom Duerig.
\newblock Scaling up visual and vision-language representation learning with
  noisy text supervision.
\newblock In \emph{International conference on machine learning}, pages
  4904--4916. PMLR, 2021.

\bibitem[Kim et~al.(2023)Kim, Park, Lee, Park, and Sohn]{language-free}
Dahye Kim, Jungin Park, Jiyoung Lee, Seongheon Park, and Kwanghoon Sohn.
\newblock Language-free training for zero-shot video grounding.
\newblock In \emph{Proceedings of the IEEE/CVF Winter Conference on
  Applications of Computer Vision}, pages 2539--2548, 2023.

\bibitem[Krishna et~al.(2017)Krishna, Hata, Ren, Fei-Fei, and
  Carlos~Niebles]{ActivityNet-Caption}
Ranjay Krishna, Kenji Hata, Frederic Ren, Li~Fei-Fei, and Juan Carlos~Niebles.
\newblock Dense-captioning events in videos.
\newblock In \emph{ICCV}, pages 706--715, 2017.

\bibitem[Li et~al.(2022)Li, Xie, Qian, Zhu, Tang, Wu, Yang, Zhuang, and
  Wang]{visa}
Juncheng Li, Junlin Xie, Long Qian, Linchao Zhu, Siliang Tang, Fei Wu, Yi~Yang,
  Yueting Zhuang, and Xin~Eric Wang.
\newblock Compositional temporal grounding with structured variational
  cross-graph correspondence learning.
\newblock In \emph{CVPR}, pages 3032--3041, 2022.

\bibitem[Li et~al.(2023)Li, Gan, Lin, Lin, Liu, Liu, and Wang]{li2023lavender}
Linjie Li, Zhe Gan, Kevin Lin, Chung-Ching Lin, Zicheng Liu, Ce~Liu, and Lijuan
  Wang.
\newblock Lavender: Unifying video-language understanding as masked language
  modeling.
\newblock In \emph{CVPR}, pages 23119--23129, 2023.

\bibitem[Liu et~al.(2022{\natexlab{a}})Liu, Qu, Di, Cheng, Xu, and Zhou]{mgsl}
Daizong Liu, Xiaoye Qu, Xing Di, Yu~Cheng, Zichuan Xu, and Pan Zhou.
\newblock Memory-guided semantic learning network for temporal sentence
  grounding.
\newblock \emph{arXiv preprint arXiv:2201.00454}, 2022{\natexlab{a}}.

\bibitem[Liu et~al.(2022{\natexlab{b}})Liu, Qu, Wang, Di, Zou, Cheng, Xu, and
  Zhou]{dscnet}
Daizong Liu, Xiaoye Qu, Yinzhen Wang, Xing Di, Kai Zou, Yu~Cheng, Zichuan Xu,
  and Pan Zhou.
\newblock Unsupervised temporal video grounding with deep semantic clustering.
\newblock In \emph{Proceedings of the AAAI Conference on Artificial
  Intelligence}, volume~36, pages 1683--1691, 2022{\natexlab{b}}.

\bibitem[Luo et~al.(2023)Luo, Huang, Gong, Jin, and Liu]{VDI}
Dezhao Luo, Jiabo Huang, Shaogang Gong, Hailin Jin, and Yang Liu.
\newblock Towards generalisable video moment retrieval: Visual-dynamic
  injection to image-text pre-training.
\newblock In \emph{CVPR}, pages 23045--23055, 2023.

\bibitem[Luo et~al.(2022)Luo, Ji, Zhong, Chen, Lei, Duan, and
  Li]{luo2022clip4clip}
Huaishao Luo, Lei Ji, Ming Zhong, Yang Chen, Wen Lei, Nan Duan, and Tianrui Li.
\newblock Clip4clip: An empirical study of clip for end to end video clip
  retrieval and captioning.
\newblock \emph{Neurocomputing}, 508:\penalty0 293--304, 2022.

\bibitem[Mun et~al.(2020)Mun, Cho, and Han]{lgi2020}
Jonghwan Mun, Minsu Cho, and Bohyung Han.
\newblock Local-global video-text interactions for temporal grounding.
\newblock In \emph{CVPR}, pages 10810--10819, 2020.

\bibitem[Nam et~al.(2021)Nam, Ahn, Kang, Ha, and Choi]{psvl}
Jinwoo Nam, Daechul Ahn, Dongyeop Kang, Seong~Jong Ha, and Jonghyun Choi.
\newblock Zero-shot natural language video localization.
\newblock In \emph{Proceedings of the IEEE/CVF International Conference on
  Computer Vision}, pages 1470--1479, 2021.

\bibitem[Radford et~al.(2021)Radford, Kim, Hallacy, Ramesh, Goh, Agarwal,
  Sastry, Askell, Mishkin, Clark, et~al.]{clip}
Alec Radford, Jong~Wook Kim, Chris Hallacy, Aditya Ramesh, Gabriel Goh,
  Sandhini Agarwal, Girish Sastry, Amanda Askell, Pamela Mishkin, Jack Clark,
  et~al.
\newblock Learning transferable visual models from natural language
  supervision.
\newblock In \emph{International conference on machine learning}, pages
  8748--8763. PMLR, 2021.

\bibitem[Regneri et~al.(2013)Regneri, Rohrbach, Wetzel, Thater, Schiele, and
  Pinkal]{Tacos}
Michaela Regneri, Marcus Rohrbach, Dominikus Wetzel, Stefan Thater, Bernt
  Schiele, and Manfred Pinkal.
\newblock Grounding action descriptions in videos.
\newblock \emph{Transactions of the Association for Computational Linguistics},
  1:\penalty0 25--36, 2013.
\newblock \doi{10.1162/tacl_a_00207}.
\newblock URL \url{https://aclanthology.org/Q13-1003}.

\bibitem[Rohrbach et~al.(2012)Rohrbach, Regneri, Andriluka, Amin, Pinkal, and
  Schiele]{tacos-from}
Marcus Rohrbach, Michaela Regneri, Mykhaylo Andriluka, Sikandar Amin, Manfred
  Pinkal, and Bernt Schiele.
\newblock Script data for attribute-based recognition of composite activities.
\newblock In \emph{Computer Vision--ECCV 2012: 12th European Conference on
  Computer Vision, Florence, Italy, October 7-13, 2012, Proceedings, Part I
  12}, pages 144--157. Springer, 2012.

\bibitem[Sanh et~al.(2019)Sanh, Debut, Chaumond, and Wolf]{DistilBERT}
Victor Sanh, Lysandre Debut, Julien Chaumond, and Thomas Wolf.
\newblock Distilbert, a distilled version of bert: smaller, faster, cheaper and
  lighter.
\newblock \emph{arXiv preprint arXiv:1910.01108}, 2019.

\bibitem[Sigurdsson et~al.(2016)Sigurdsson, Varol, Wang, Farhadi, Laptev, and
  Gupta]{sigurdsson2016hollywood}
Gunnar~A Sigurdsson, G{\"u}l Varol, Xiaolong Wang, Ali Farhadi, Ivan Laptev,
  and Abhinav Gupta.
\newblock Hollywood in homes: Crowdsourcing data collection for activity
  understanding.
\newblock In \emph{ECCV}, pages 510--526. Springer, 2016.

\bibitem[Tran et~al.(2015)Tran, Bourdev, Fergus, Torresani, and Paluri]{c3d}
Du~Tran, Lubomir Bourdev, Rob Fergus, Lorenzo Torresani, and Manohar Paluri.
\newblock Learning spatiotemporal features with 3d convolutional networks.
\newblock In \emph{Proceedings of the IEEE international conference on computer
  vision}, pages 4489--4497, 2015.

\bibitem[Wang et~al.(2022{\natexlab{a}})Wang, Wu, Liu, and Yan]{pzvmr}
Guolong Wang, Xun Wu, Zhaoyuan Liu, and Junchi Yan.
\newblock Prompt-based zero-shot video moment retrieval.
\newblock In \emph{Proceedings of the 30th ACM International Conference on
  Multimedia}, pages 413--421, 2022{\natexlab{a}}.

\bibitem[Wang et~al.(2021{\natexlab{a}})Wang, Xing, and
  Liu]{wang2021actionclip}
Mengmeng Wang, Jiazheng Xing, and Yong Liu.
\newblock Actionclip: A new paradigm for video action recognition.
\newblock \emph{arXiv preprint arXiv:2109.08472}, 2021{\natexlab{a}}.

\bibitem[Wang et~al.(2022{\natexlab{b}})Wang, Li, Li, He, Huang, Zhao, Zhang,
  Xu, Liu, Wang, et~al.]{wang2022internvideo}
Yi~Wang, Kunchang Li, Yizhuo Li, Yinan He, Bingkun Huang, Zhiyu Zhao, Hongjie
  Zhang, Jilan Xu, Yi~Liu, Zun Wang, et~al.
\newblock Internvideo: General video foundation models via generative and
  discriminative learning.
\newblock \emph{arXiv preprint arXiv:2212.03191}, 2022{\natexlab{b}}.

\bibitem[Wang et~al.(2021{\natexlab{b}})Wang, Chen, and Jiang]{vca}
Zheng Wang, Jingjing Chen, and Yu-Gang Jiang.
\newblock Visual co-occurrence alignment learning for weakly-supervised video
  moment retrieval.
\newblock In \emph{ACM MM}, pages 1459--1468, 2021{\natexlab{b}}.

\bibitem[Wang et~al.(2022{\natexlab{c}})Wang, Wang, Wu, Li, and Wu]{mmn}
Zhenzhi Wang, Limin Wang, Tao Wu, Tianhao Li, and Gangshan Wu.
\newblock Negative sample matters: A renaissance of metric learning for
  temporal grounding.
\newblock In \emph{AAAI}, volume~36, pages 2613--2623, 2022{\natexlab{c}}.

\bibitem[Wu et~al.(2020)Wu, Li, Liu, and Lin]{tsp}
Jie Wu, Guanbin Li, Si~Liu, and Liang Lin.
\newblock Tree-structured policy based progressive reinforcement learning for
  temporally language grounding in video.
\newblock In \emph{AAAI}, volume~34, pages 12386--12393, 2020.

\bibitem[Xu et~al.(2019)Xu, He, Plummer, Sigal, Sclaroff, and Saenko]{qspn}
Huijuan Xu, Kun He, Bryan~A Plummer, Leonid Sigal, Stan Sclaroff, and Kate
  Saenko.
\newblock Multilevel language and vision integration for text-to-clip
  retrieval.
\newblock In \emph{AAAI}, volume~33, pages 9062--9069, 2019.

\bibitem[Yang et~al.(2021)Yang, Feng, Ji, Wang, and Chua]{dcm}
Xun Yang, Fuli Feng, Wei Ji, Meng Wang, and Tat-Seng Chua.
\newblock Deconfounded video moment retrieval with causal intervention.
\newblock In \emph{Proceedings of the 44th international ACM SIGIR conference
  on research and development in information retrieval}, pages 1--10, 2021.

\bibitem[Zhang et~al.(2020{\natexlab{a}})Zhang, Sun, Jing, and Zhou]{vslnet}
Hao Zhang, Aixin Sun, Wei Jing, and Joey~Tianyi Zhou.
\newblock Span-based localizing network for natural language video
  localization.
\newblock In \emph{Proceedings of the 58th annual meeting of the association
  for computational linguistics}, pages 6543--6554, Online, July
  2020{\natexlab{a}}. Association for Computational Linguistics.
\newblock URL \url{https://www.aclweb.org/anthology/2020.acl-main.585}.

\bibitem[Zhang et~al.(2020{\natexlab{b}})Zhang, Peng, Fu, and Luo]{2dtan}
Songyang Zhang, Houwen Peng, Jianlong Fu, and Jiebo Luo.
\newblock Learning 2d temporal adjacent networks for moment localization with
  natural language.
\newblock In \emph{AAAI}, volume~34, pages 12870--12877, 2020{\natexlab{b}}.

\bibitem[Zheng et~al.(2022{\natexlab{a}})Zheng, Huang, Chen, and Liu]{cnm}
Minghang Zheng, Yanjie Huang, Qingchao Chen, and Yang Liu.
\newblock Weakly supervised video moment localization with contrastive negative
  sample mining.
\newblock In \emph{Proceedings of the AAAI Conference on Artificial
  Intelligence}, volume~36, pages 3517--3525, 2022{\natexlab{a}}.

\bibitem[Zheng et~al.(2022{\natexlab{b}})Zheng, Huang, Chen, Peng, and
  Liu]{CPL}
Minghang Zheng, Yanjie Huang, Qingchao Chen, Yuxin Peng, and Yang Liu.
\newblock Weakly supervised temporal sentence grounding with gaussian-based
  contrastive proposal learning.
\newblock In \emph{Proceedings of the IEEE/CVF Conference on Computer Vision
  and Pattern Recognition (CVPR)}, 2022{\natexlab{b}}.

\end{thebibliography}
}

\end{document}